\documentclass[letterpaper, 10 pt, conference]{ieeeconf}

\IEEEoverridecommandlockouts                              

\usepackage{multicol}
\usepackage{bm}
\usepackage{url}
\usepackage{graphicx} 
\usepackage{newtxtext,newtxmath}      
\usepackage[mathscr]{euscript}  
\usepackage{amsmath} 

\usepackage{amssymb}  
\usepackage{flushend}
\usepackage{multirow}
\usepackage{siunitx}
\usepackage{xcolor}
\usepackage{adjustbox}
\usepackage{array}
\usepackage{multirow}
\usepackage{booktabs}
\let\labelindent\relax
\usepackage[inline]{enumitem}
\usepackage[hang,flushmargin]{footmisc}
\usepackage{microtype}
\usepackage{amsmath}
\usepackage{threeparttable}
\usepackage[resetlabels]{multibib}
\usepackage{cite}

\newcites{New}{References}

\newcommand{\secref}[1]{Sec.~\ref{#1}}
\renewcommand{\eqref}[1]{Eq.~(\ref{#1})}
\newcommand{\figref}[1]{Fig.~\ref{#1}}

\newcolumntype{P}[1]{>{\centering\arraybackslash}p{#1}}

\newcommand{\motionnet}{MAPT$_\text{Motion}$}
\newcommand{\appearancenet}{MAPT$_\text{Appearance}$}

\newcommand{\netname}{MAPT }

\usepackage{pifont}
\usepackage{color,array}

\usepackage[pdfencoding=auto, colorlinks=true]{hyperref} 

\makeatletter
\def\zapcolorreset{\let\reset@color\relax\ignorespaces}
\def\colorrows#1{\noalign{\aftergroup\zapcolorreset#1}\ignorespaces}
\makeatother

\title{\LARGE \bf 
Learning Appearance and Motion Cues for Panoptic Tracking
}

\author{Juana Valeria Hurtado, Sajad Marvi, Rohit Mohan, and Abhinav Valada
\thanks{Department of Computer Science, University of Freiburg, Germany.}%
\thanks{This work was funded by the German Research Foundation (DFG) Emmy Noether Program grant number 468878300.}}

\begin{document}

\maketitle
\thispagestyle{empty}
\pagestyle{empty}

\raggedbottom

\begin{abstract}


Panoptic tracking enables pixel-level scene interpretation of videos by integrating instance tracking in panoptic segmentation. This provides robots with a spatio-temporal understanding of the environment, an essential attribute for their operation in dynamic environments.
In this paper, we propose a novel approach for panoptic tracking that simultaneously captures general semantic information and instance-specific appearance and motion features.
Unlike existing methods that overlook dynamic scene attributes, our approach leverages both appearance and motion cues through dedicated network heads. These interconnected heads employ multi-scale deformable convolutions that reason about scene motion offsets with semantic context and motion-enhanced appearance features to learn tracking embeddings.
Furthermore, we introduce a novel two-step fusion module that integrates the outputs from both heads by first matching instances from the current time step with propagated instances from previous time steps and subsequently refines associations using motion-enhanced appearance embeddings, improving robustness in challenging scenarios.
Extensive evaluations of our proposed \netname model on two benchmark datasets demonstrate that it achieves state-of-the-art performance in panoptic tracking accuracy, surpassing prior methods in maintaining object identities over time. To facilitate future research, we make the code available at \mbox{\url{http://panoptictracking.cs.uni-freiburg.de}}.
\end{abstract}

\section{Introduction}

Semantic interpretation of dynamic scenes is a fundamental aspect of robot perception, as it enables robots to comprehend and safely navigate their environment~\cite{kappeler2024few,gosala2023skyeye}. Panoptic tracking~\cite{hurtado2020mopt} facilitates the perception of such scenes by incorporating pixel-level semantic segmentation and instance segmentation with temporally associated object IDs across video frames. Recent works propose end-to-end learning methods for panoptic tracking that primarily combine panoptic segmentation and multi-object tracking architectures~\cite{hurtado2020mopt,weber2021step,wanghvps}. These detection-based methods first learn visual features that encode the appearance characteristics of objects in the scene and then compare them across frames with the aim of finding similarities and assigning a unique tracking ID to each object throughout the sequence. Despite the success of such methods, their performance remains limited in conditions such as complex lighting or texture changes and occlusions. 
For example, appearance cues are usually misleading when distinguishing between objects that look similar and are insufficient to match objects that are partially or completely occluded in one of the frames. 
Conversely, research on Multi-Object Tracking and Segmentation (MOTS) has shown that motion features are invaluable for learning to track objects. As a result, various MOTS methods employ a propagation-based approach, which includes a mask propagation architecture that exploits previous frame cues to predict the object's displacement and shape at the current time~\cite{bertasius2020classifying, cheng2021modular, miao2022self}.

\begin{figure}
    \centering
    \includegraphics[width=0.85\linewidth]{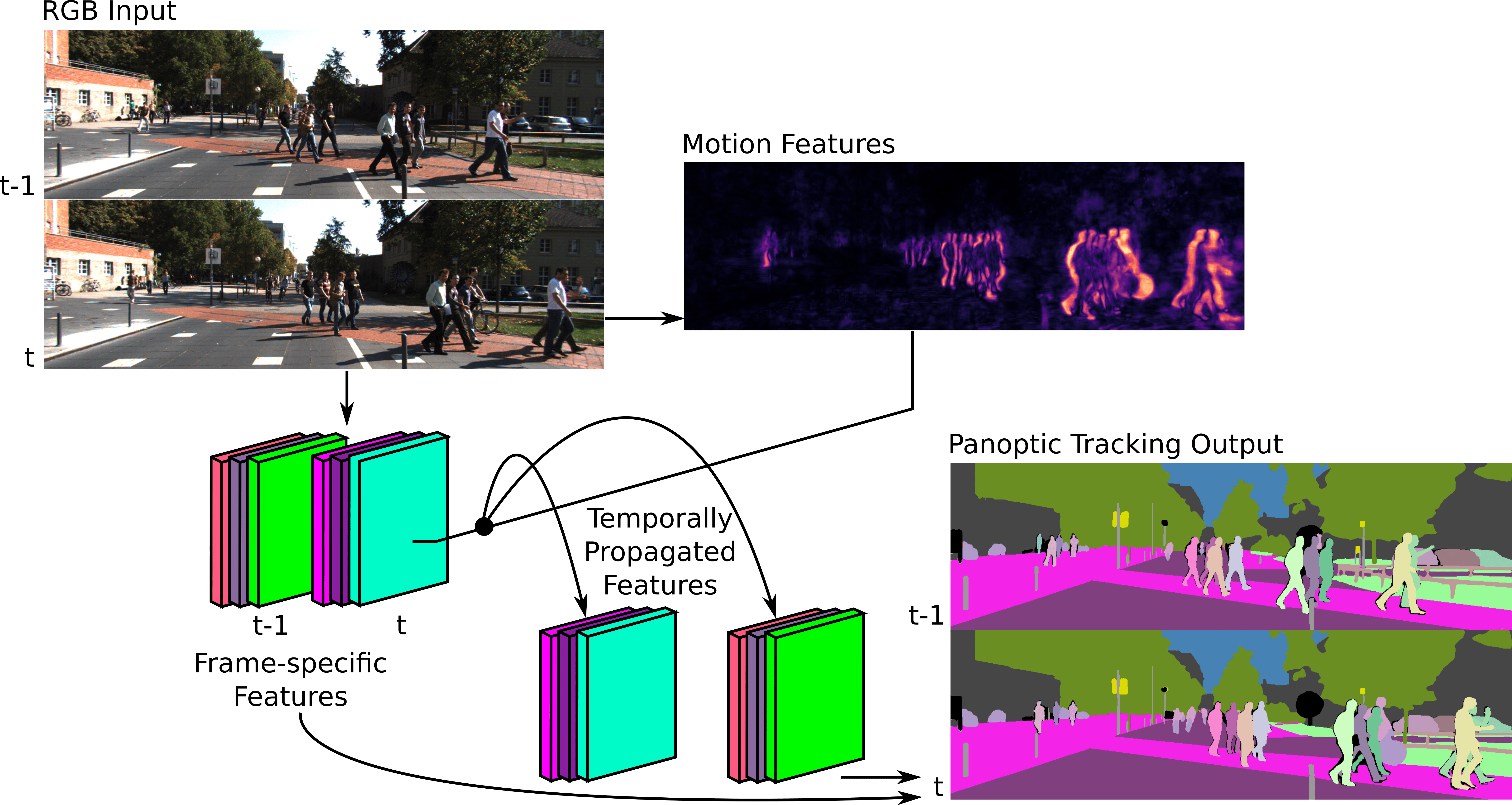}
    \caption{Proposed MAPT framework integrating appearance and motion cues for panoptic tracking. Our approach improves tracking robustness by leveraging both appearance and motion features, enabling better spatio-temporal scene understanding in dynamic environments.}
    \label{fig:teaser}
\end{figure}


In this direction, some panoptic tracking approaches incorporate motion information using external flow information to mitigate tracking errors due to visual changes~\cite{weber2021step, wanghvps}. These methods rely on training an additional optical flow prediction network. Subsequently, they propagate object masks from the current time step into adjacent frames. However, since the motion and flow models are trained disentangled from the panoptic tracking network, they ignore temporal dynamics such as motion direction and patterns. Additionally, optical flow networks are typically computationally expensive and excessively large to incorporate into end-to-end panoptic tracking approaches. Unlike MOTS methods, which integrate motion cues when learning objects' temporal coherence. Consequently, these MOTS methods are able to adapt to appearance changes, deformations, and fast movement of objects. Nevertheless, these rich motion cues have not yet been exploited to learn panoptic tracking.


With this in mind, we propose the novel Motion and Appearance-aware Panoptic Tracking (MAPT) architecture that simultaneously leverages cues from image sequences at two levels of abstraction: appearance-based and motion-aware, as we illustrate in \figref{fig:teaser}. \netname holistically learns general scene semantics, appearance-specific object features, and motion features. It consists of a shared backbone and interconnected heads that leverage spatio-temporal video features. We propose \motionnet~that uses encoded semantic feature changes between frames and predicted object masks to propagate dynamic object masks to adjacent frames. Additionally, we introduce \appearancenet, which learns tracking embeddings using propagated features from the \motionnet~to enhance object appearance features. Furthermore, we propose a fusion module that encodes learned scene appearance and motion cues to generate a coherent panoptic tracking output. We perform extensive evaluations on the KITTI-STEP and MOTChallenge-STEP~\cite{weber2021step} datasets that show that our proposed \netname achieves state-of-the-art performance using the Panoptic And Tracking (PAT) metric with less computational overhead than panoptic tracking methods that use external optical flow models.


We summarize our main contributions as follows:
\begin{enumerate}
    \item \netname that leverages complementary appearance and motion cues for panoptic tracking.
    \item \motionnet, a motion-aware mask propagation head that reasons about intermediate semantic features from different time steps for mask propagation.
    \item \appearancenet, an appearance-aware tracking head that leverages the propagated features to enhance the visual features with motion cues.
    \item A novel fusion block that effectively integrates panoptic segmentation and instance tracking predictions using tracking outputs from motion and appearance heads. 
    \item Extensive benchmarking results on two datasets and ablation studies that demonstrate the utility of incorporating both appearance and motion cues.
    \item Publicly available code at \url{http://panoptictracking.cs.uni-freiburg.de}.
\end{enumerate}

\section{Related Work}


{\parskip=2pt
\noindent\textit{Panoptic Tracking}: 
Panoptic tracking methods can be grouped into two categories: top-down and bottom-up. Top-down approaches use a generalized feature extractor and multiple heads for semantic segmentation, instance segmentation, and tracking~\cite{hurtado2020mopt, kim2020video}. The instance segmentation and tracking heads generate bounding boxes, masks, and tracking embedding vectors from proposal features. In parallel, the semantic head predicts the segmentation mask, and the outputs of the different heads are fused, associating objects across frames based on their appearance embeddings. Alternative approaches exploit motion features for panoptic tracking by integrating an external flow network to propagate instance masks onto an adjacent time step~\cite{kim2022tubeformer} or wrap instance segmentation features~\cite{wanghvps}.
Bottom-up methods first predict the semantic segmentation mask and find object centers in adjacent frames to identify and track instances that have connected features across frames~\cite{weber2021step,mohan2022perceiving,vodisch2023codeps}. This approach requires extensive post-processing and is impractical for online tracking.} 

More recently, end-to-end learning methods, such as~\cite{li2022video,cheng2021mask2former}, simultaneously learn segmentation and tracking with a query-based approach using transformers. They use dynamic kernels for panoptic segmentation that encode the appearance features of objects and scene context from different time frames. The authors in \cite{li2022video} use these kernels to learn their respective embedding vectors that are fed into a data association algorithm. Although this unified approach facilitates leveraging scene context by learning features from sequential images, they have limited motion information, therefore relying more on association algorithms. Recently, Video-KMaX~\cite{shin2023video} extended the kernel-based approach by introducing a video instance segmentation model that jointly learns panoptic segmentation and tracking across video frames. This model learns instance-specific kernel representations to efficiently segment and track objects across time without relying on explicit optical flow estimation. In contrast, our proposed MAPT learns both motion and appearance features, allowing for specialized predictions that can support performance in scenarios where object appearance is insufficient.

{\parskip=2pt
\noindent\textit{Appearence-based MOTS}: 
Unlike panoptic tracking, MOTS only tracks instances in a video. Several works have proposed a unified end-to-end network to address MOTS based on the appearance of objects~\cite{voigtlaender2019mots, luiten2019video, luiten2020unovost, wang2019fast, porzi2020learning}.  These approaches learn to map object-specific features onto an embedding space, which is then used to match objects throughout the sequence~\cite{yang2019video,zhou2020tracking, wang2020towards} by computing associations between their respective instances using data association algorithms such as the Hungarian method~\cite{he2017mask} and QDTracker~\cite{pang2021quasi}. Despite notable advancements, this framework faces significant challenges. For example, it can be challenging to track objects with similar semantics and appearance, associate objects across varying lighting and weather conditions, and deal with occlusions. These challenges may lead to erroneous object associations. The authors in \cite{yang2021self} show that variations in illumination, weather conditions, or abrupt modifications in the shape of an object can affect an object's visible features, making it challenging to associate objects accurately.} 

{\parskip=2pt
\noindent\textit{Motion-aware and Propagation-based MOTS}:
MOTS can also be addressed using mask propagation-based methods. These techniques propagate instance masks from previous time steps onto the current frame to identify the position and shape of objects in the current frame. They typically represent objects as spatio-temporal connections between pixels using instance masks from previous time steps and their motion information. There are two primary methods to achieve this objective. The first uses scene flow-based propagation of instances~\cite{cheng2018fast}, while the second exploits scene difference values and deformable convolutions to learn how to project an instance mask onto the current frame~\cite{jiang2019video}. By incorporating both appearance and motion features in their learning algorithms, these methods have been observed to perform effectively in scenarios where appearance features alone are insufficient to distinguish target objects from other instances or background pixels~\cite{yang2018efficient}.} 
\section{Technical Approach}
\label{fig:architecture}

\begin{figure*}
    \centering
    \includegraphics[width=0.85\linewidth]{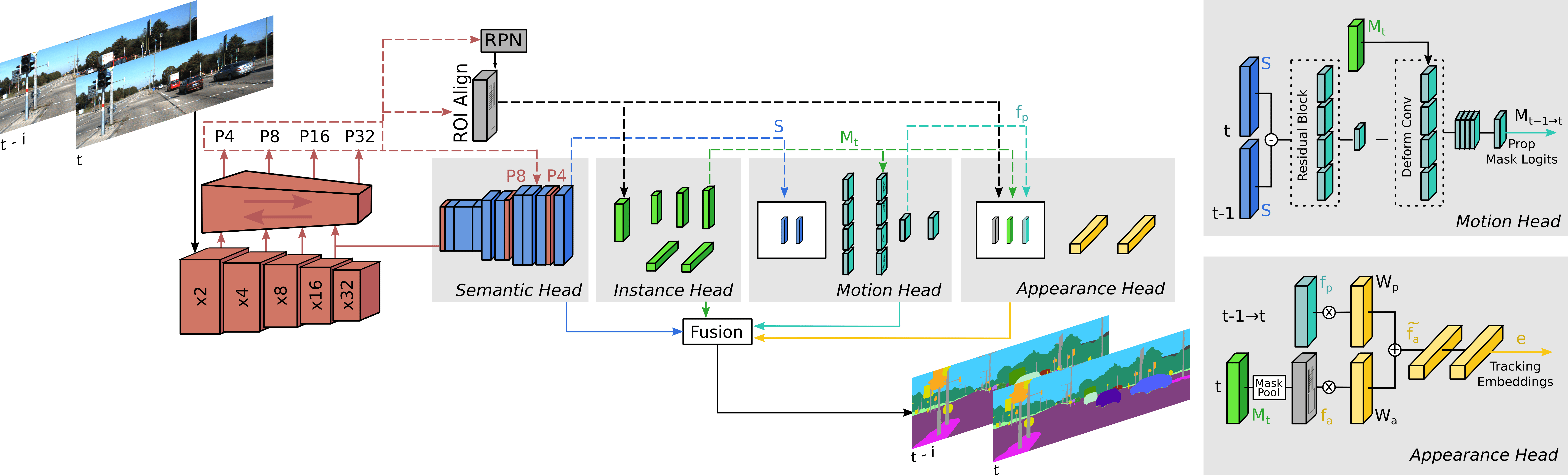}
    \caption{Our \netname architecture is composed of a shared backbone and four interconnected heads for semantic segmentation, instance segmentation, and motion-based and appearance-based object tracking. Our motion head employs multi-scale deformable convolutions to capture rich semantic and motion features, while the appearance head focuses on learning instance-specific visual representations. These two heads complement each other, as motion cues help in scenarios where appearance alone is ambiguous, while appearance features provide stability when motion is unreliable. By combining the outputs with our fusion block, enhance the robustness of panoptic tracking in dynamic environments.}
    \label{fig:arch}
    \vspace{-0.5cm}
\end{figure*}

Our framework, as shown in Fig.~\ref{fig:arch}, aims to assign a semantic label $c^{ij}_t$ to each pixel in an image $I_t$ of size $H \times W$ in a video of length $T$, where $i \in [\![1,W]\!]$, $j \in [\![1,H]\!]$, and $t \in [1,T]$. We also generate consistent instance IDs ($ID_t$) for dynamic objects using visual and motion features ($f_a$ and $f_p$). Our \netname architecture consists of a shared backbone, connected to a semantic segmentation head and an RPN with ROI align layers for task-specific heads (instance, tracking, and mask propagation). These heads utilize enhanced features and generate specific predictions fused for a temporally coherent output at the input resolution. We divide the network into four sections: panoptic segmentation, appearance network, motion network, and a fusion module.

\subsection{Panoptic Segmentation}

We base the panoptic segmentation network on the APSNet~\cite{mohan2022amodal} architecture. We use RegNetY-8.0GF~\cite{radosavovic2020designing} as the shared backbone with a 2-way Feature Pyramid Network (FPN) to extract features at multiple scales. While aggregating multi-scale features, the encoder facilitates bidirectional information flow to obtain $P_x$ features across four resolutions, each with 256 filters. Each $P_x$ represents a downsampled size of  $\times 1/4, \times 1/8, \times 1/16, \times 1/32$. 
We input $P_x$ into the Region Proposal Network (RPN) to generate bounding box candidates. Then, RoIAlign extracts features of each candidate to feed the instance segmentation head, appearance head \appearancenet~(\secref{sec:track_head}), and motion head \motionnet~(\secref{sec:motion_head}).
The instance segmentation head simultaneously predicts the class, bounding box, and segmentation mask of each instance candidate using two parallel branches: one for mask segmentation and one for bounding box regression and classification. We use a modified Mask R-CNN with standard convolutions, batch normalization, and ReLU activation layers replaced with depthwise separable convolution, iABN sync, and Leaky ReLU, respectively. We optimize the instance head by minimizing the sum of the losses corresponding to each part of the instance head as
\begin{align}\label{eq:instance_loss}
    L_{\text{instance}}(d, \hat{d}) =& L_{\text{os}} + L_{\text{op}} + L_{\text{cls}} + L_{\text{bbx}} + L_{\text{mask}},
\end{align}
where $d, \hat{d}$ indicate ground truth and prediction for each instance detection and $L_{\text{os}}$, $L_{\text{op}}$, $L_{\text{cls}}$, $L_{\text{bbx}}$, and $L_{\text{mask}}$ are the object score, proposal, classification, bounding box, and mask segmentation losses respectively, as defined in \cite{he2017mask}. 
 
The semantic segmentation head aims to increase context aggregation, expanding the receptive field. Therefore, it uses $P4$ and $P8$ outputs of the FPN and features from the shared backbone as inputs. The semantic segmentation head first takes the $\times 1/16$ downsampled features from the fourth encoder stage. It passes them through successive dilation convolutions and a subsequent Dense Prediction Cell (DPC)  module that processes the dilated block output \cite{chen2018searching}. The following upsampling operations generate features that we concatenate to $P8$ and $P4$ to include information at multiple resolutions to obtain the semantic features $S$. We train the semantic segmentation head with weighted per-pixel log-loss~\cite{bulo2017loss} given by
\begin{equation}
L_{semantic}(\Theta) = \dfrac{1}{|N|}\sum_{n \in N} \sum_{ij}w_{ij}(P_{ij} \cdot \log \hat{P_{ij}}),
\label{eq:semantic_loss}
\end{equation}
\begin{align}
 w_{ij}=
   \begin{cases}
         \dfrac{4}{WH}, & \text{if pixel (i,j)}  \in \text{25\% worst prediction }\\
         0, & \text{otherwise}
      \end{cases},
\end{align}
where $N$ is the batch set, $P_{ij}$ is the ground truth for a given image in the batch set, $\hat{P_{ij}}$ is the predicted probability for the pixel $(i,j)$ assigned to class $c\in P$.

\subsection{{Motion-aware Tracking Head}}
\label{sec:motion_head}

Our \motionnet~propagates the semantic features $f_{t-}(d) \rightarrow{}\hat{f}_t(d)$ of a detected instance $d$ from frame $({I_{t-1}})$ into a reference frame $(I_t)$. Therefore, the goal is to use the propagated features $\hat{f}_t(d)$ to obtain a corresponding segmentation mask at the reference frame $M(d)^{pred}_{t - 1 \rightarrow{} t}$. To do so, our \motionnet~takes as inputs the frame semantic features $S_{t-1}$ and $S_{t}$ from the semantic head and the mask $M(d)_t$ from the instance head. Similar to \cite{bertasius2020classifying}, \motionnet~comprises a motion and residual block, a set of deformable convolutions, and a $1 \times 1$ convolution layer.

In contrast to \cite{bertasius2020classifying}, which uses the backbone features $P$ as input, we add semantic context by using $S$ instead, exploiting panoptic tracking information. Therefore, our motion block encodes the semantic feature changes between the two frames by computing the difference between their semantic features $S_{t-1} - S_{t}$. Subsequently, the residual block predicts four motion offsets obtained with convolution layers, each with dilation $[1,3,6,12]$. The motion offsets and the instance masks $M(d)_{t-1}$ are the input to a set of $k \times k$ deformable convolutions. We concatenate the outputs from the deformable convolutions to account for different instance sizes, obtaining the propagated features $\hat{f}(d)_t = f(d)_{t-1 \rightarrow{} t}$. Thereafter, a $1 \times 1$ convolution layer outputs instance masks for each object  $M_{_{t - 1 \rightarrow{} t}}$.
We compute the loss by comparing the IoUs of the propagated mask and the masks in the reference frame as follows:
\begin{align}
    L_{motion} = \sum_d \sum_t 1 - sIoU\{M_{_{t}}^{gt} (d), M_{_{t - 1 \rightarrow{} t}}^{pred}(d)\}
\end{align}
through all $D$ instances and $T$ frames.

\subsection{Appearance-aware Tracking Head}
\label{sec:track_head}
We leverage the visual aspect of objects by learning embeddings $e(d)_t$ from appearance features $\Tilde{f_a}(d)_t$ of a set of bounding box proposals ${D_t}$. In addition to the bounding box features, we include information from the instance and motion head to enhance the features in both the spatial and temporal domains. Having the unfiltered proposal output from the RPN and their respective ROIAling features, we obtain the set of bounding box proposals $D_t$ by randomly sampling $128$ bounding boxes so that $25\%$ are positive. A proposed bounding box is positive if the IoU between it and the bounding box's ground truth is greater than $0.5$. Our \appearancenet~uses object shape information $M(d)_{t}$ obtained in the instance segmentation head. Therefore, by means of a mask pooling operator, we ignore features that are outside the object's mask if the corresponding bounding box is positive, obtaining $f_a(d)_t$. Additionally, we use the propagated features $\hat{f_p}(d)_t$ from the \motionnet~to learn motion-enhanced appearance features as follows:
\begin{align}
    \Tilde{f_a}(d)_t = W_a*f_a(d)_t + W_p*\hat{f}_p(d)_t,
\end{align}
where $W_a$ and $W_b$ are adaptable weights that we learn through attention. The rest of the \appearancenet~consists of two fully connected layers that encode the enhanced features embedding vectors $e(d)_t$. We compute the matching embedding loss as the cross entropy loss based on the predicted similarity logits of a pair of images. We compute the similarity with the cosine distance 
\begin{align}
    \delta &=  \dfrac {e(d)_t \cdot e(d)_{t-\tau}} {\left\| e(d)_{t}\right\| _{2}\left\| e(d)_{t-\tau}\right\| _{2}}, \\
    L_{appearance} &= -\sum_{track_{gt}=1}^{N_{tracks}} y_{[d_t,d_{t-\tau}],track_{gt}}\log(\delta_{o,track_{gt}}),
\end{align}
where $N_{tracks}$ is the number of trackIDs and $y$ is $0$ or $1$ indicating if the embeddings $(e(d)_{t},e(d)_{t-\tau})$ belong to the same $track_{gt}$.

\subsection{Fusion Module}
The panoptic fusion module combines semantic and instance head logits based on mask confidence scores, including both 'thing' and 'stuff' classes. We obtain the Fused Logits ($FL$) as:
\begin{align}
    FL = (\sigma(ML_A) + \sigma(ML_B)) \odot (ML_A + ML_B),
\end{align}
where $ML_A$ and $ML_B$ are the Mask Logits ($ML$) from the semantic and instance heads, respectively. Subsequently, our motion and appearance assignment involves a two-stage strategy. Initially, we assign identical instance IDs to propagated masks and current masks that are situated in close proximity to the motion offset. This is achieved through a motion offset-based process. Second, we refine the temporal association of instances by utilizing appearance embeddings in critical situations. We determine the cosine similarities of the detection embeddings when a propagated mask is not nearby. If the similarity is greater than a specific threshold and the semantic class of the detections is the same, the detection is assigned to a track.

\subsection{{Implementation Details}}
We first pre-train a panoptic segmentation network on the Cityscapes \cite{cordts2016cityscapes} dataset. We perform this pretraining using the classes in common with Cityscapes for each respective model. These pretrained networks are then fine-tuned on each dataset. We train our model with SGD with a momentum of $0.9$ using a multi-step learning rate schedule. We initialize a base learning rate and train the model for a specific number of iterations. Then, we decrease the learning rate by a factor of 10 at each milestone and continue training for 10 epochs. We use an initial learning rate $lr_{base}$ of 0.01. Initializing the training, we have a warm-up phase where the $lr_{base}$ is increased linearly from $\frac{1}{3}\cdot lr_{base}$ to $lr_{base}$ in 200 iterations. We use the rest of the hyperparameters as specified in \cite{mohan2022amodal}. We use the PyTorch deep learning library for implementing our approach, and we trained the panoptic tracking models on a system with 8 Nvidia GeForce RTX 3090 (24GB) GPUs and the AMD EPYC™ 7452 CPU at 3.20 GHz with a total of 128 cores. We train \netname end-to-end by minimizing the final loss and adding all the individual losses.

\section{Experimental Evaluation}
\label{sec:experiments}

\begin{table*}
\centering
\footnotesize
\caption{Panoptic tracking results on the KITTI-STEP validation set.}
\vspace{-0.2cm}
\begin{threeparttable}
\begin{tabular}{p{4.8cm}|P{0.9cm}P{0.9cm}P{0.9cm}|P{0.9cm}P{0.9cm}P{0.9cm}}
\toprule
Network & PAT & PQ & TQ & STQ & AQ & SQ \\
  & (\%) & (\%) & (\%) & (\%) & (\%) & (\%) \\
\noalign{\smallskip}\hline\hline\noalign{\smallskip}
Mask2former (siain~\cite{ryu2021end}) & $44.79$ & $44.64$ & $44.94$ & $52.94$ & $46.73$ & $58.15$ \\
Motion-DeepLab~\cite{weber2021step} & $54.05$ & $52.20$ & $56.03$ & $60.71$ & $54.28$ & $67.90$ \\
MOPT~\cite{hurtado2020mopt} & $50.99$ & $46.43$ & $56.55$ & $62.71$ & $57.00$ & $69.01$ \\
VPS*~\cite{kim2020video} & $49.05$ & $44.78$ & $54.22$ & $49.51$ & $44.71$ & $54.83$ \\
VPS~\cite{kim2020video} & $-$ & $-$ & $-$ & $56.00$ & $52.00$ & $61.00$ \\
TubeFormer-DeepLab~\cite{kim2022tubeformer} & $-$ & $-$ & $-$ & $65.00$ & $61.00$ & $62.00$ \\
IMTNet~\cite{wang2022instance} & $-$ & $-$ & $-$ & $62.00$ & $56.00$ & $70.00$ \\
Video K-Net~\cite{li2022video} & $61.29$ & $\underline{56.66}$ & $66.74$ & ${73.33}$ & $\underline{71.95}$ & ${74.74}$ \\
Video-kMaX~\cite{shin2023video} & $\underline{62.16}$ & $55.20$ & $\underline{70.01}$ & $\mathbf{79.00}$ & $\mathbf{78.80}$ & $\underline{78.90}$ \\
\midrule
\netname (ours) & $\mathbf{64.20}$ & $\mathbf{59.04}$ & $\mathbf{70.33}$ & $\underline{75.38}$ & $69.47$ & $\mathbf{81.80}$ \\
\bottomrule
\end{tabular}
\label{tab:kittistepval}
The final results for \netname are computed with scale augmentation during testing, outperforming the baselines by $2.04\%$ in the PAT score. The baselines marked with * were reimplemented to obtain PAT scores. 
\end{threeparttable}
\end{table*}

\begin{table}
\centering
\footnotesize
\caption{Panoptic tracking results on KITTI-STEP test set.}
\vspace{-0.2cm}
\begin{threeparttable}
\begin{tabular}{p{3.4cm}|P{0.9cm}P{0.9cm}P{0.9cm}}
\toprule
Network & STQ & AQ & SQ \\
  & (\%) & (\%) & (\%) \\
\noalign{\smallskip}\hline\hline\noalign{\smallskip}
siain~\cite{ryu2021end} & $57.87 $ & $55.16 $ & $60.71 $\\
Motion-DeepLab~\cite{weber2021step} & $52.19 $ & $45.55 $ & $59.81 $\\
TubeFormer-DeepLab~\cite{kim2022tubeformer} & $65.25 $ & $60.59$ & $\mathbf{70.27}$ \\
Video-kMaX~\cite{shin2023video} & $\mathbf{68.47} $ & $\mathbf{67.20} $ & $\underline{69.77} $\\
\midrule
\netname (ours) & $ \underline{67.38}$ &	$\underline{66.54}$&	$68.23$ \\
\bottomrule
\end{tabular}
\label{tab:kittisteptes}
\end{threeparttable}
\end{table}

\begin{table}
\centering
\footnotesize
\caption{Panoptic tracking results on MOTChallenge-STEP test set.}
\vspace{-0.2cm}
\begin{threeparttable}
\begin{tabular}{p{3.4cm}|P{0.9cm}P{0.9cm}P{0.9cm}}
\toprule
Network & STQ & AQ & SQ \\
  & (\%) & (\%) & (\%) \\
\noalign{\smallskip}\hline\hline\noalign{\smallskip}
siain~\cite{ryu2021end} & $31.8$ & $15.4$ & $65.7$\\
EffPS$_MM$ ~\cite{mohan2021efficientps} & $42.8$ & $26.4$ & $\mathbf{69.2}$\\
IPL$_ETRI$~\cite{wanghvps} & $\mathbf{48.6}$ & $\mathbf{43.3}$ & $54.5$ \\
\midrule
\netname (ours) & $ \underline{46.7}$ &	$\underline{32.3}$&	$\underline{67.4}$ \\
\bottomrule
\end{tabular}
\label{tab:mottest}
\end{threeparttable}
\end{table}

\begin{table*}
\centering
\footnotesize
\caption{Ablation study on the backbone and tracking heads.}
\vspace{-0.2cm}
\begin{threeparttable}
\begin{tabular}{l|l|cc|ccc|ccc}
\toprule
Network & Track Method & Motion & Appearence & PAT & PQ & TQ & STQ & AQ & SQ \\
  & & & & (\%) & (\%) & (\%) & (\%) & (\%) & (\%) \\
\noalign{\smallskip}\hline\hline\noalign{\smallskip}
Panoptic-DeepLab\cite{weber2021step} & IoUmatch & $\times$ & $\checkmark$ & $54.05$ & $52.20$ & $56.03$ & $60.71$ & $54.28$ & $67.90$\\
APSNet \cite{mohan2022amodal} & IoUmatch & $\times$ & $\checkmark$ & $59.45$ & $56.91$ & $62.22$ & $63.58$ & $59.22$ & $68.26$ \\
APSNet \cite{mohan2022amodal} & FlowNet2\cite{ilg2017flownet} & $\checkmark$ & $\times$ &  $61.01$ & $56.91$ & $65.76$ & $ 66.16$ & $64.12$ & $68.26 $ \\
APSNet \cite{mohan2022amodal} & MaskProp~\cite{bertasius2020classifying}& $\checkmark$ & $\times$ & $59.59$ & $56.91$ & $62.54$ & $65.58$ & $63.46$ & $68.26$ \\
\appearancenet (ours) & Appearance (ours) & $\times$ & $\checkmark$& $59.84$ & $56.91$ & $63.10$ & $65.58$ & $63.02$ & $68.26$ \\
\motionnet (ours) & Motion (ours) & $\checkmark$ & $\times$ & $61.61$ & $56.91$ & $67.16$ & $67.32$ & $67.32$ & $68.26$ \\
\midrule
\netname (ours) & Motion$\&$Appearance (ours) & $\checkmark$ & $\checkmark$ & $\mathbf{62.65}$ & $\mathbf{58.00}$ & $\mathbf{68.13}$ & $\mathbf{70.91}$ & $\mathbf{68.80}$ & $\mathbf{73.10}$ \\
\bottomrule
\end{tabular}
\label{tab:ablationAll}
This ablation study confirms that combining motion and appearance cues improves performance across all metrics, demonstrating their complementary benefits for panoptic tracking.
\end{threeparttable}
\vspace{-0.3cm}
\end{table*}

In this section, we present extensive evaluations and ablation studies to demonstrate the novelty of our contributions.

\subsection{Metrics and Datasets}
\label{sec:datasets}
Different metrics have been proposed to evaluate panoptic tracking. The most recent metrics are Panoptic And Tracking (PAT)~\cite{fong2022panoptic} and Segmentation and Tracking Quality (STQ)~\cite{weber2021step}.Both combine segmentation and tracking terms. STQ, designed for pixel-level evaluation, follows five principles: \textit{i)} Pixel-level analysis, \textit{ii)} No threshold-based matching, \textit{iii)} No penalty for mistake correction, \textit{iv)} Precision-recall association, and \textit{v)} Error decoupling.
The STQ is calculated using the geometric mean formula: $\text{STQ} = \sqrt{\text{AQ} \times \text{SQ}}$. AQ measures the quality of track ID assignment with a weighted IoU metric, while SQ evaluates semantic segmentation quality by class with a standard IoU metric. 
STQ identifies objects and assigns semantic classes based on pixels, while PAT is instance-focused and sensitive to ID changes.

Although PAT was proposed initially for LiDAR panoptic tracking, we adapted this metric to our image-based task, defined as the harmonic mean $\text{PAT} = (2 \times \text{PQ} \times \text{TQ})/(\text{PQ} + \text{TQ})   \in [0,1],$
where PQ is standard panoptic quality~\cite{kirillov2019panoptic} and TQ is the tracking quality \cite{fong2022panoptic}, defined as: 
\begin{align}
TQ(g) &= \sum([1-\frac{IDS(g)}{N_{IDS}(g)}]\times AS(g))^{\frac{1}{2}} , \label{eq:track_quality_1} \\
AS(g) &= \frac{1}{|g|}\sum_{p,|p \cap g|\ne 0}TPA(p,g)\times IoU_{a}(p,g).
\end{align}

The Association Score ($AS$) is calculated using true positive association ($TPA$) between ground truth tracks ($gt_{tracks}$) and any other object prediction with an ID ($p$) that has a mask overlap greater than 0.5 IoU. $AS$ maintains consistent tracking and penalizes the removal of correctly segmented instances. $TQ(g)$ penalizes track fragmentation for high $g$ tracks using ID switch rates. $N_{IDS}$ is the max ID switches over track length $|g|$. ID switch occurs when two consecutive frames of instance $g$ have inconsistent IDs or no matching predictions.
Finally, TQ is computed as follows:
\begin{align}
TQ &= \frac{1}{gt_{tracks}}\sum_{g\in gt_{tracks}} TQ(g), \in [0,1] \label{eq:track_quality_2},
\end{align}
where $gt_{tracks}$ corresponds to a group of unique instance IDs.

The KITTI-Segmenting and Tracking Every Pixel (KITTI-STEP) dataset is based on KITTI-MOTS, which includes tracking IDs for pedestrians and cars - the most important moving objects~\cite{weber2021step}. However, KITTI-STEP goes further by adding annotations for various "stuff" categories such as buildings and vegetation, while still maintaining tracking IDs for "thing" classes throughout the entire sequence. This dataset contains 19 classes, including 2 "thing" classes and 17 "stuff" classes, as in the Cityscapes dataset. There are 21 sequences available for training and 29 for testing. MOTChallenge-STEP, similar to KITTI-STEP, is built on top of the MOTS-Challenge~\cite{voigtlaender2019mots} dataset. It provides tracking instance annotations for pedestrians as well as pixel-wise annotations of semantic categories in urban street views. This dataset contains two sequences for training and two for testing, with 7 semantic classes divided into 1 "thing" class and 6 "stuff" classes.

\subsection{Quantitative Results}

We evaluate \netname across multiple benchmarks, comparing it to state-of-the-art panoptic tracking methods. Specifically, Tab.~\ref{tab:kittistepval} presents its performance on the KITTI-STEP validation set. Along with STQ, we report PAT and its separable terms (TQ, PQ), which emphasize instance-level tracking performance. We observe that \netname achieves the highest PAT score of $64.20\%$, surpassing Video-kMaX~\cite{shin2023video} by $2.04\%$. This indicates improved instance-level consistency and fewer identity switches, which is crucial for tracking dynamic objects in real-world applications. Although Video-kMaX achieves a higher STQ score ($79.00\%$ vs. $75.38\%$), \netname outperforms it in both TQ (70.33\% vs. 70.01\%) and PQ ($59.04\%$ vs. $55.20\%$). Since STQ primarily evaluates pixel-wise segmentation accuracy, it does not fully reflect object-level tracking quality. The improvements in PAT, TQ, and PQ indicates that \netname maintains stronger object identity coherence across frames, reducing fragmentation and identity switches. These results demonstrate that \netname provides more reliable instance tracking, making it a robust solution for panoptic tracking in dynamic environments.

Tab.~\ref{tab:kittisteptes} and Tab.~\ref{tab:mottest} report MAPT’s performance on the KITTI-STEP and MOTChallenge-STEP test sets, respectively. On KITTI-STEP, MAPT achieves a STQ score of $67.38\%$, comparable to Video-kMaX~\cite{shin2023video} which achieves $68.47\%$ STQ. Since the benchmark only evaluates STQ, it does not reflect instance identification performance. Given the higher PAT scores on the validation set, MAPT is likely to offer improved identity consistency across frames.

\begin{figure}
\footnotesize
\centering
\setlength{\tabcolsep}{0.03cm}
\begin{tabular}{p{0.30cm}p{2.70cm}p{2.70cm}p{2.70cm}}
  & $t$ &$t+\tau_1$ &$t+\tau_2$ \\
      \rotatebox{90}{RGB} &
      \includegraphics[width=\linewidth]{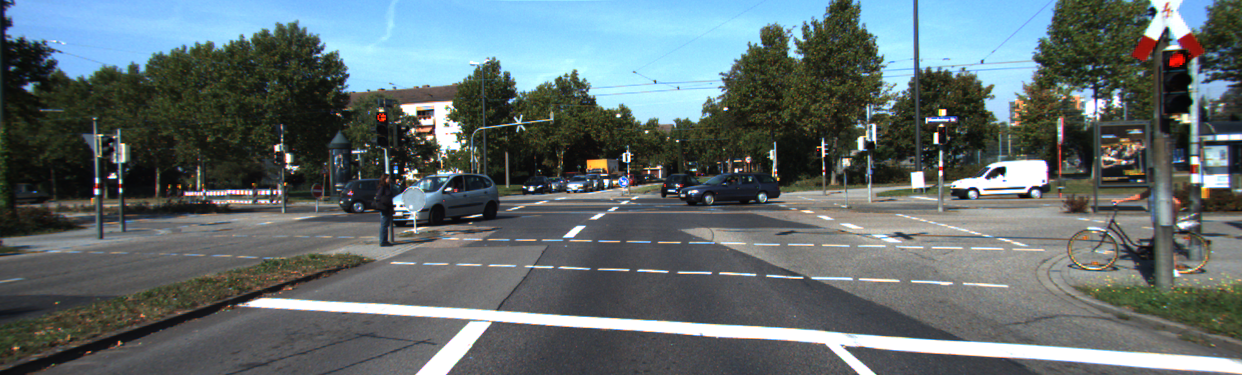} &
      \includegraphics[width=\linewidth]{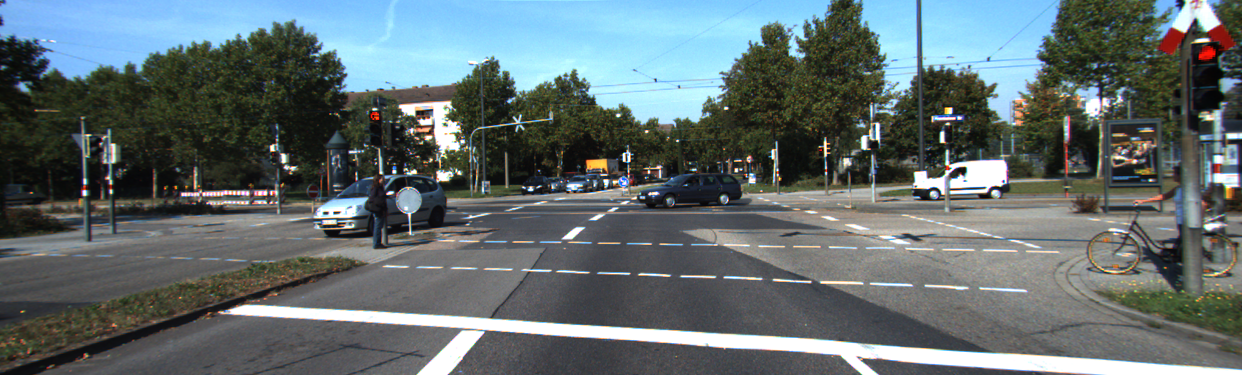} &
      \includegraphics[width=\linewidth]{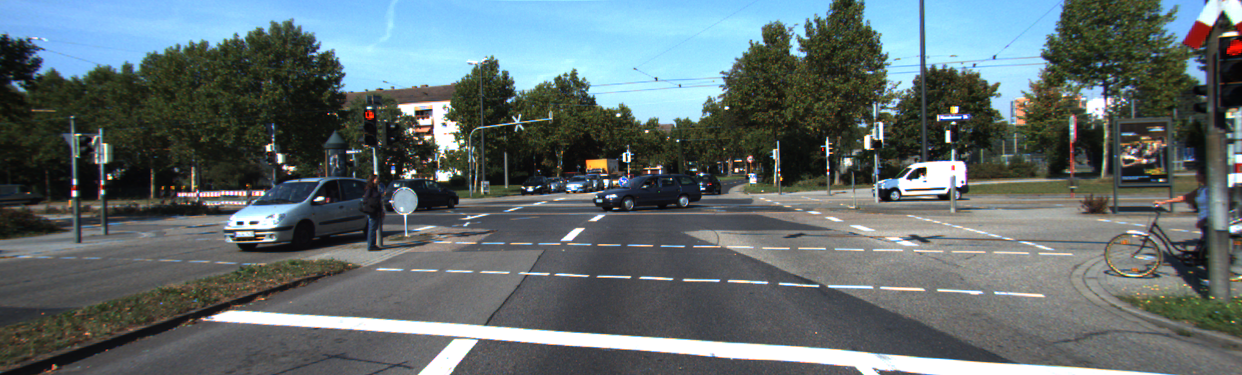} \\

      \rotatebox{90}{MAPT$_M$} &
      \includegraphics[width=\linewidth]{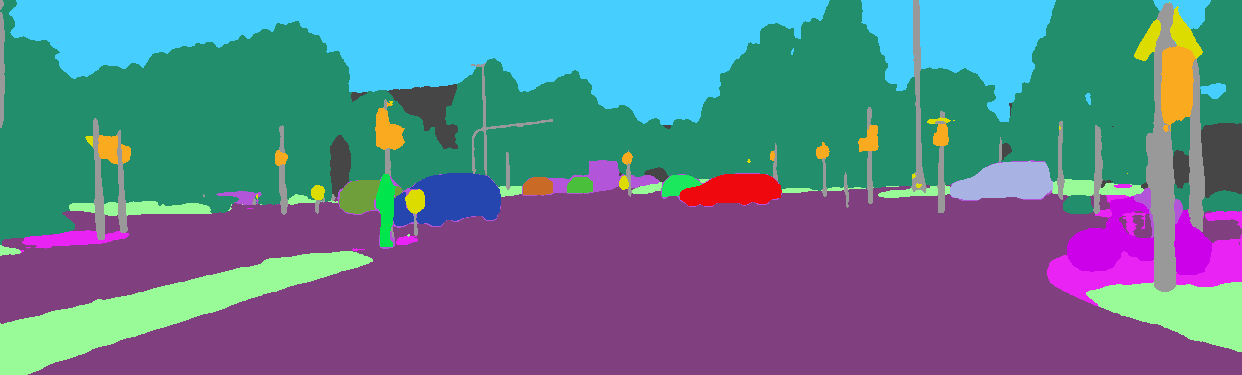} &
      \includegraphics[width=\linewidth]{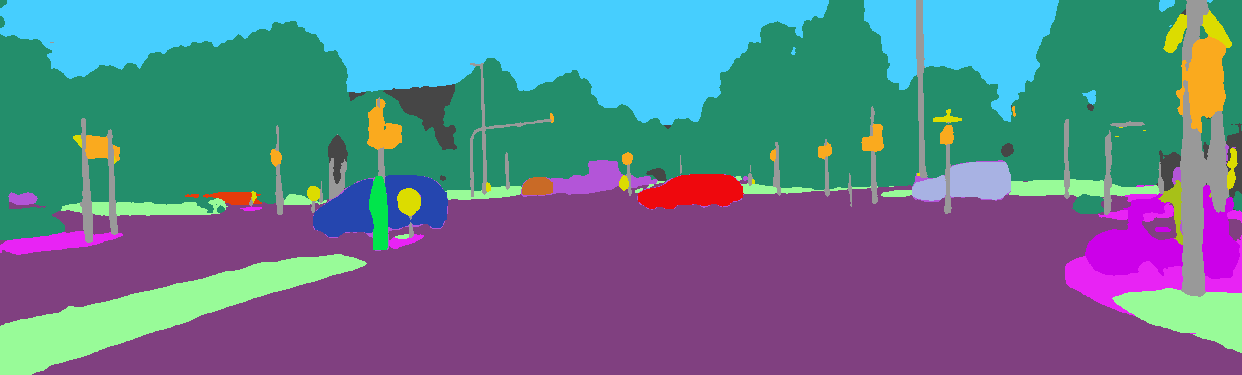} &
      \includegraphics[width=\linewidth]{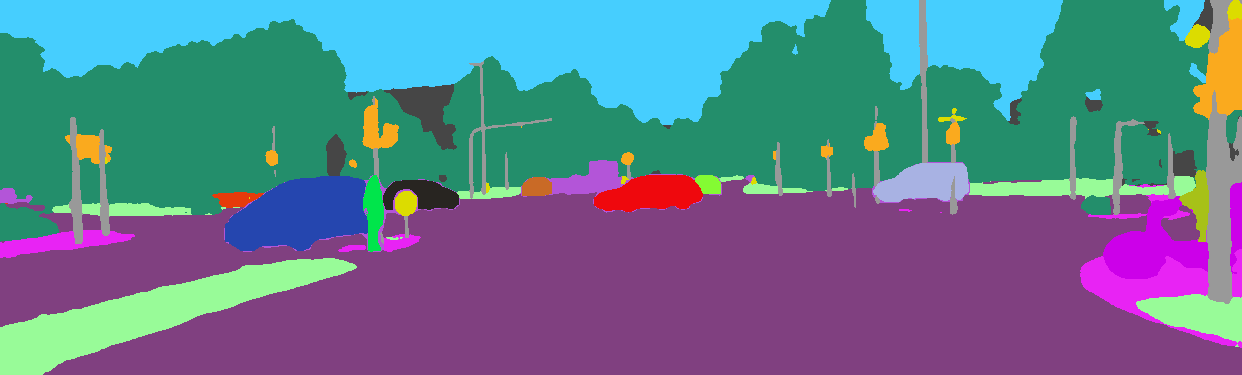} \\
      
      \rotatebox{90}{MAPT$_A$} &
      \includegraphics[width=\linewidth]{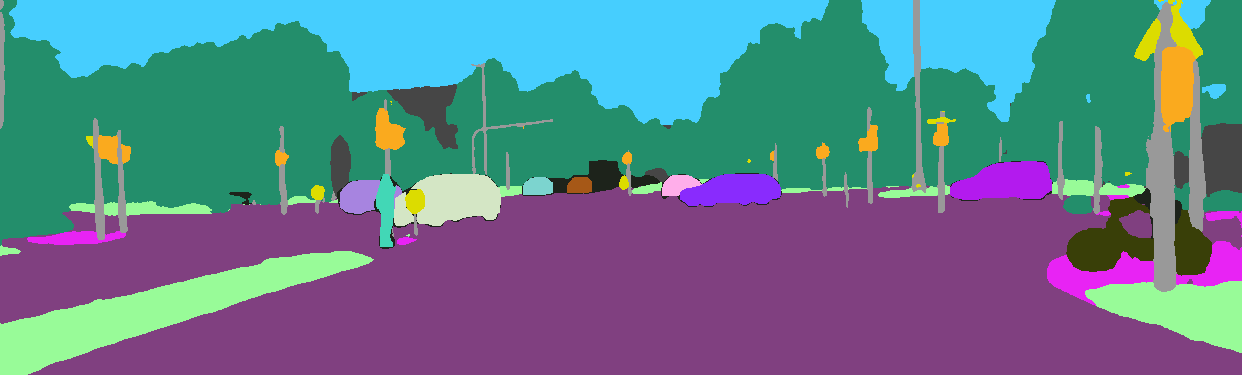} &
      \includegraphics[width=\linewidth]{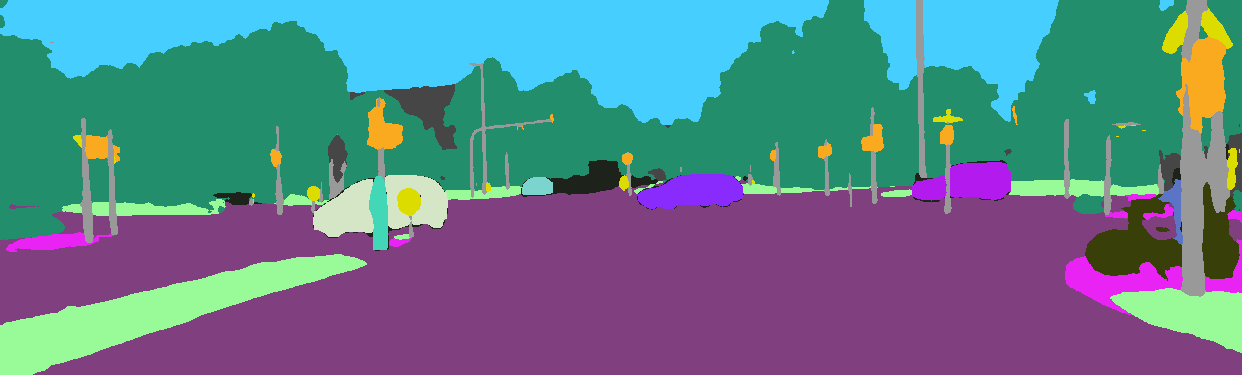} &
      \includegraphics[width=\linewidth]{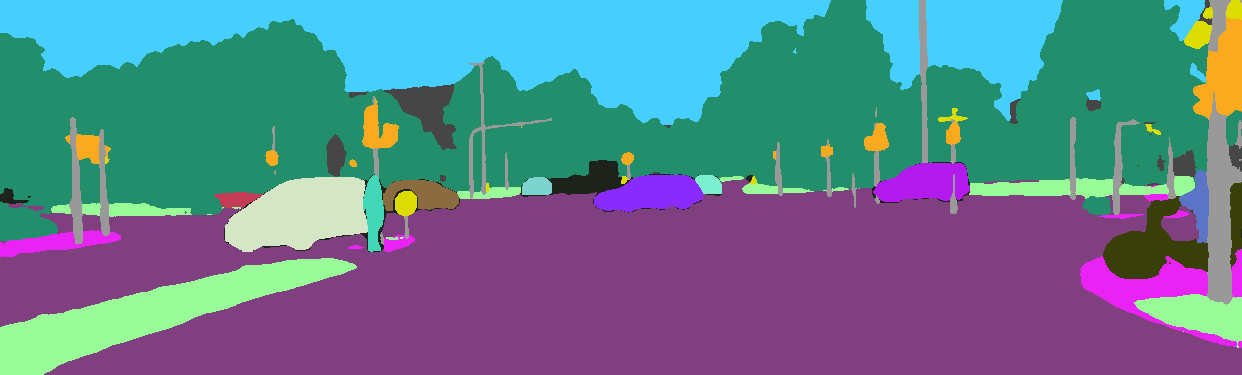} \\

      \rotatebox{90}{Video-knet} &
      \includegraphics[width=\linewidth]{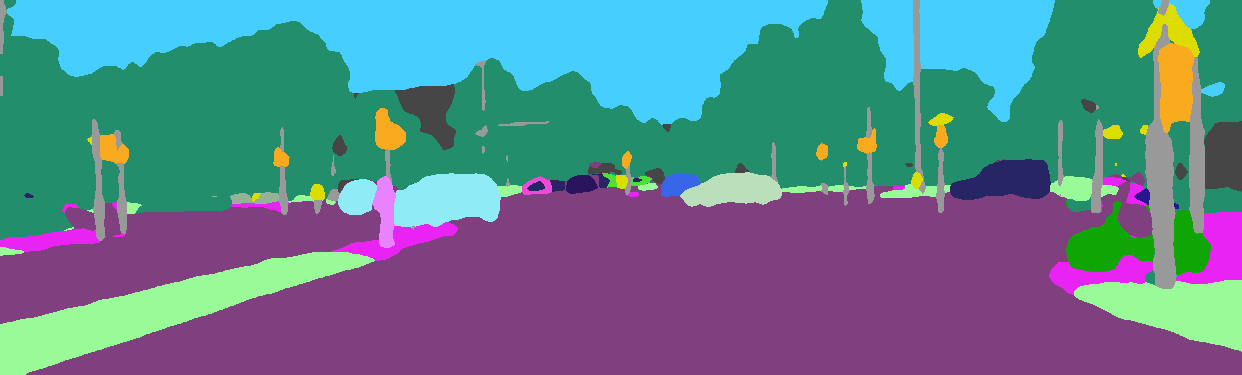} &
      \includegraphics[width=\linewidth]{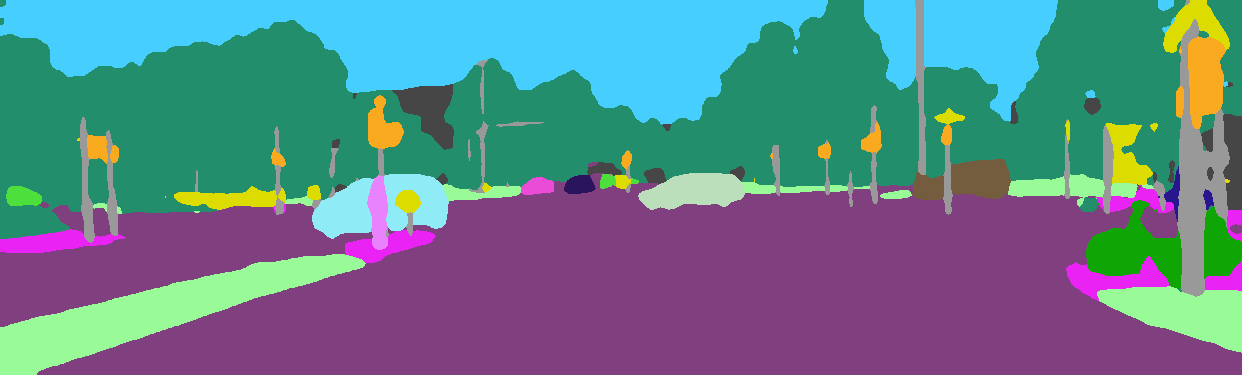} &
      \includegraphics[width=\linewidth]{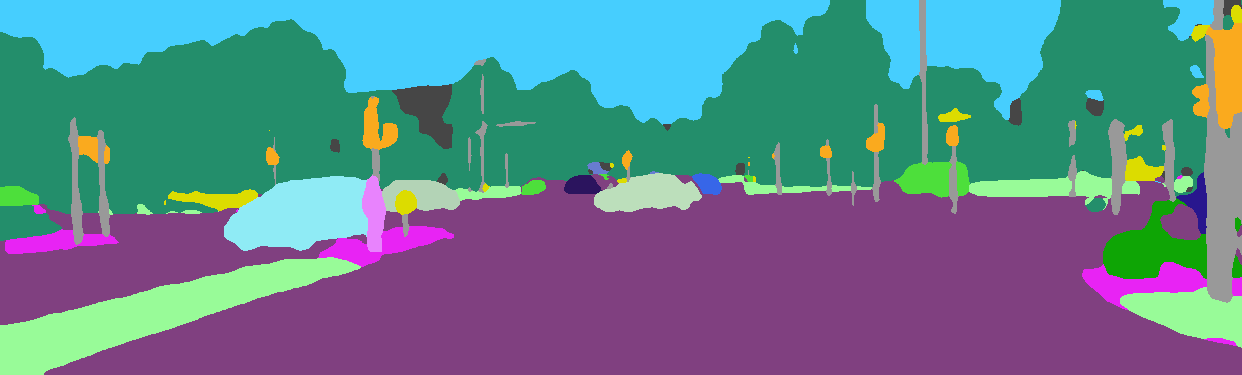} \\
\end{tabular}
\caption{Qualitative comparisons of panoptic tracking from our proposed MAPT  (first and second rows) with Video-Knet (last row). Each column shows a different frame in a video. These results demonstrate that our MAPT method effectively maintains object identity and shape consistency through occlusions.\looseness=-1}
\label{fig:qualitative}
\end{figure}

On MOTChallenge-STEP, \netname achieves a STQ score of $46.7\%$, with a segmentation quality of $67.4\%$, which is comparable to state-of-the-art methods. While IPL$_ETRI$~\cite{wanghvps} reports a higher overall STQ, our model demonstrates strong segmentation consistency despite the dataset’s inherent constraints—only two test sequences and a primary focus on pedestrian tracking. Although \netname does not achieve state-of-the-art performance on this dataset, it offers a well-balanced trade-off between segmentation accuracy and tracking stability, reinforcing its robustness across diverse benchmarks. Notably, this dataset lacks a validation set, and the benchmark does not evaluate PAT. Therefore, it excludes the influence of instance tracking performance, making it difficult to assess our model’s strength in tracking consistency. Given \netname demonstrates advantages in maintaining object identities on other datasets, its full potential may not be fully reflected in this benchmark.

\subsection{Ablation Study}

In Tab.~\ref{tab:ablationAll}, we present ablation results supporting the final architectural configuration of MAPT. In the base model, we use Panoptic-Deeplab for panoptic segmentation with IouMatching for tracking as presented in \cite{weber2021step}. By leveraging our APSNet as the base network, we achieve gains in PAT and STQ of $5.4\%$ and $2.87\%$, respectively. Subsequently, we use FlowNet~\cite{ilg2017flownet} to propagate the mask of dynamic objects and match them based on the IoU between propagated masks and the current masks to obtain tracking results. We then integrate a mask propagation network into the panoptic tracking framework MaskProp~\cite{bertasius2020classifying}. Nevertheless, incorporating this network does not outperform using Flownet. Therefore, we improve the performance by incorporating semantic features into the mask propagation head. Finally, adding the fusion module, which includes the appearance tracking embeddings, we obtain the best performance of $62.65\%$ in PAT score and $70.91\%$ in STQ.

We present additional ablations in Tab.~\ref{tab:motionablation} and Tab.~\ref{tab:appareanceablation}. For motion, we show we achieve better performance using segmentation features and with less computational complexity with a reduction of GFlops from $993.13$ to $3.50$, compared to using an external optical flow network. For appearance, we present results using panoptic segmentation predictions and trackers such as ByteTrack. We show that learning appearance features to obtain tracking embeddings increases overall performance. Specifically, by enhancing the RoIAling features with the masks obtained from the instance head and with the motion features from the motion head, we obtain the best configuration for \appearancenet.

\begin{table*}
\caption{Ablation study on \motionnet, comparing computational complexity and performance.}
\vspace{-2mm}
\centering
\footnotesize
\begin{tabular}{p{2.3cm}|p{1.2cm}p{1.2cm}|p{0.9cm}p{0.9cm}p{0.9cm}|p{0.9cm}p{0.9cm}p{0.9cm}|p{0.9cm}p{0.9cm}}
\toprule
Network & Flow & Features & PAT & PQ & TQ & STQ & AQ & SQ & Params. & FLOPs \\
 &  & & (\%) & (\%) & (\%) & (\%) & (\%) & (\%) & (M) & (G) \\
\noalign{\smallskip}\hline\hline\noalign{\smallskip}
\motionnet$^1$ & \checkmark & $\times$  & $61.01$ & $56.91$ & $65.76$ & $66.16$ & $64.12$ & $68.26$ & $162.51$ & $993.13$ \\
\motionnet$^2$ & $\times$ & Backbone  & $59.59$ & $56.91$ & $62.54$ & $65.58$ & $63.46$ & $68.26$ & $0.38$ & $3.50$ \\
\midrule
\motionnet & $\times$ & Seg  & $\mathbf{61.61}$ & $\mathbf{56.91}$ & $\mathbf{67.16}$ & $\mathbf{67.79}$ & $\mathbf{67.32}$ & $\mathbf{68.26}$ & $\mathbf{0.38}$ & $\mathbf{3.50}$ \\
\bottomrule
\end{tabular}
\label{tab:motionablation}

\vspace{2mm}

\makebox[\linewidth]{\parbox{0.93\linewidth}{\raggedright
Ablations on \motionnet\ assess the impact of external optical flow models and segmentation features on efficiency and tracking. Adding segmentation improves motion-based tracking while reducing computational cost.
}}

\end{table*}

\begin{table*}
\caption{Ablation Study on Appearance-Based Panoptic Tracking.}
\vspace{-2mm}
\centering
\footnotesize
\begin{tabular}{p{2.3cm}|P{2.0cm}P{2.0cm}|P{0.9cm}P{0.9cm}P{0.9cm}|P{0.9cm}P{0.9cm}P{0.9cm}}
\toprule
Network & Features  & Tracker & PAT & PQ & TQ & STQ & AQ & SQ \\
  &   & & (\%) & (\%) & (\%) & (\%) & (\%) & (\%) \\
\noalign{\smallskip}\hline\hline\noalign{\smallskip}
\appearancenet$^1$ & $\times$ & OC-SORT  & $45.96$ & $56.91$ & $38.55$ & $50.65$ & $37.59$ & $68.26$ \\
\appearancenet$^2$ & $\times$ & ByteTrack & $51.04$ & $56.91$ & $46.28$ & $54.80$ & $44.00$ & $68.26$ \\
\appearancenet$^3$ & $\times$ & Ioumatch & $45.91$ & $56.35$ & $38.73$ & $51.65$ & $39.07$ & $68.26$ \\
\appearancenet$^4$  & Bbox ROIAlign  & Embed Dist & $54.75$ & $55.73$ & $53.80$ & $62.98$ & $57.91$ & $68.26$ \\
\appearancenet$^5$  & Mask $\star$&   Embed Dist & $59.78$ & $56.91$ & $61.43$ & $65.00$ & $61.91$ & $68.26$ \\
\midrule
\appearancenet & Mask + Motion $\star$ &   Embed Dist & $\mathbf{59.84}$ & $\mathbf{56.91}$ & $\mathbf{63.10}$ & $\mathbf{65.58}$ & $\mathbf{63.02}$ & $\mathbf{68.26}$ \\
\bottomrule
\end{tabular}
\label{tab:appareanceablation}
\vspace{2mm}

\makebox[\linewidth]{\parbox{0.93\linewidth}{\raggedright
We analyze object trackers and appearance features for panoptic tracking, showing that ROIAlign with mask logits and motion cues ($\star$) improves embedding quality and tracking performance.
}}
\vspace{-0.3cm}
\end{table*}

\subsection{Qualitative Results}
\begin{figure}[t]
\footnotesize
\centering
\setlength{\tabcolsep}{0.03cm}
    \centering
    \includegraphics[width=0.9\linewidth]{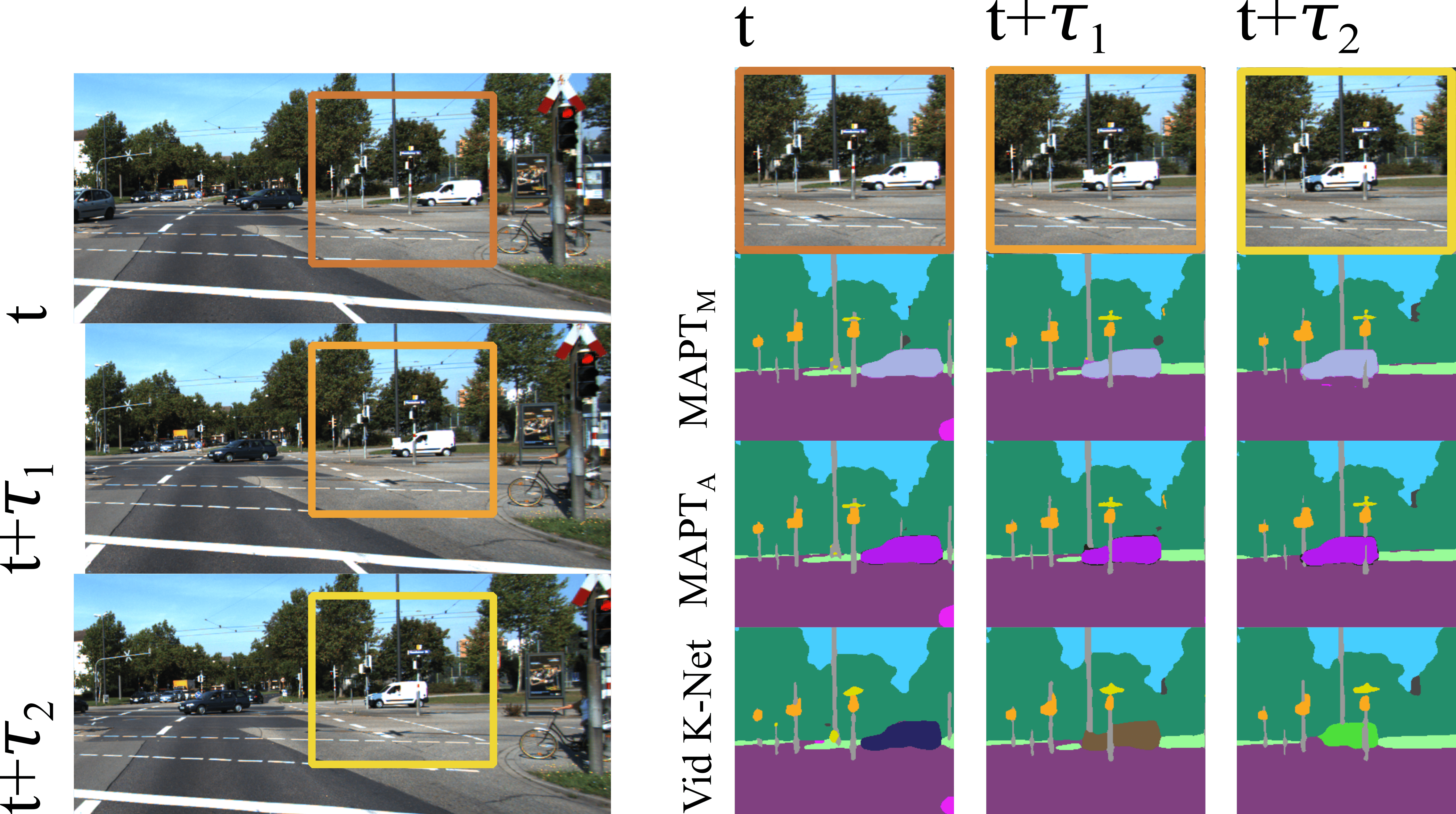}
    \caption{
We compared our proposed MAPT model with Video-Knet by analyzing three video frames, highlighting how occlusion-related visual distortion affects prediction accuracy. These results highlight that our MAPT model better preserves track identity and object shape under occlusions, leading to more consistent panoptic tracking}
    \label{fig:qualitative2}
    \vspace{-0.3cm}
\end{figure}

In Fig.~\ref{fig:qualitative}, we visualize an example of a sequence of predictions. This scene is particularly challenging as it contains multiple dynamic objects that become partially occluded. We observe that Video-Knet fails to track the car entering the scene on the right, depicted with a purple color in frame $t$. This car is partially occluded in frame $t+\tau_1$, causing a switch in the track ID now depicted in brown color. Moreover, the shape of the car masks is also affected. In the following frame ($t+\tau_2$), the mask deformation is more noticeable, with part of the car segmented as road, and the trackID switches once more to green. In contrast, \netname is able to track this car correctly. Our appearance-only network also fails to detect and track the cars in the back. As a result, our motion network shows the best performance in this scene. Furthermore, Fig.~\ref{fig:qualitative2} illustrates with more detail how our \netname can accurately maintain track ID and object shape even when the car is partially occluded.

\begin{figure}[t]
\footnotesize
\centering
\setlength{\tabcolsep}{0.03cm}
    \centering
    \includegraphics[width=0.75\linewidth]{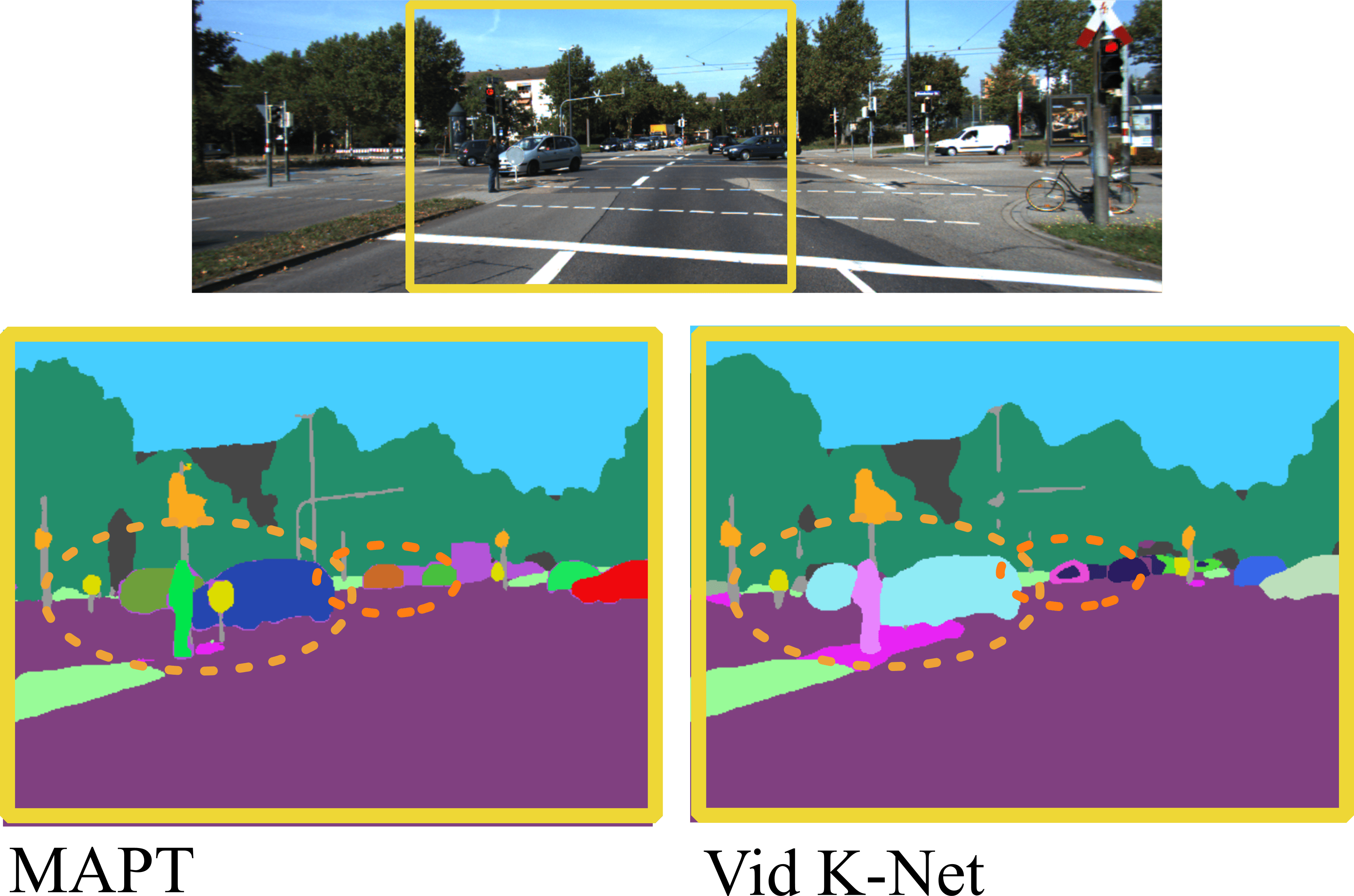}
    \caption{Comparison between our proposed MAPT and Video-Knet based on close-up analysis of three frames in a video presenting visual distortion due to occlusion that affects the predictions. These observations highlight the advantage of our MAPT method in handling occlusions and preserving object integrity.}
    \label{fig:qualitative3}
\end{figure}

Fig.~\ref{fig:qualitative3} illustrates a zoomed-up portion of the image at frame $t$. We observe segmentation quality differences between the predictions of Video-Knet and \netname highlighted with dotted orange circles. Video-knet combines the predicted masks of two different cars. This segmentation is depicted in light blue and is more severely penalized with PQ than with SQ. Similarly, a smaller segmentation on the right is presented as an unfilled pink area with Video-Knet, while in \netname is predicted as a completed area. 

\section{Conclusion}
\label{sec:conclusions}

In this work, we introduced MAPT, a novel motion- and appearance-aware panoptic tracking framework that enhances intra- and inter-frame scene understanding. By integrating panoptic segmentation with motion-aware tracking embeddings, our approach improves instance association and segmentation consistency, leading to fewer identity switches and more stable tracking. Our extensive evaluation on the KITTI-STEP validation set demonstrates that \netname achieves a $2.04\%$ improvement in the PAT score, highlighting its ability to maintain object identities more consistently than prior methods. Additionally, we analyze the impact of motion and appearance cues, showing that jointly learning both leads to a more balanced panoptic tracking solution.

Unlike methods that prioritize segmentation consistency across frames, our approach explicitly focuses on panoptic tracking by ensuring both accurate segmentation and robust instance association over time. While segmenting and tracking every pixel is important, identity switches can still occur without significantly affecting STQ performance. In contrast, \netname is designed to improve panoptic tracking performance, as reflected in its higher PAT score, demonstrating more stable instance associations and better handling of dynamic objects.
While \netname provides a competitive trade-off between segmentation accuracy and tracking robustness, several challenges remain. The choice of the temporal interval during training significantly affects both object tracking and mask propagation, making it an important factor for optimization. Future work will focus on adaptive interval selection and more efficient tracking mechanisms to enhance tracking quality while reducing computational overhead. 
Overall, our approach represents a significant step toward more reliable and interpretable panoptic tracking while significantly improving dynamic scene understanding. These advancements make \netname a promising solution for real-world applications.

\footnotesize
\bibliographystyle{IEEEtran}
\bibliography{references}

\end{document}